\documentclass[letterpaper]{article} 
\usepackage{aaai25}  
\usepackage{times}  
\usepackage{helvet}  
\usepackage{courier}  
\usepackage[hyphens]{url}  
\usepackage{graphicx} 
\usepackage{natbib}  
\usepackage{caption} 
\usepackage{algorithm}
\usepackage{algorithmic}

\usepackage[table,xcdraw]{xcolor}
\usepackage{colortbl}

\usepackage{times}
\usepackage{latexsym}
\usepackage{tabularray}
\usepackage{subfigure}
\usepackage{tcolorbox}

\newenvironment{purplebox}
{\begin{tcolorbox}[colback=purple!5,boxrule=0pt,leftrule=2pt,boxsep=2pt,arc=0pt,fontupper=\normalsize]}
{\end{tcolorbox}}

\newcommand{\tablefontsize}{\fontsize{9pt}{10pt}\selectfont}

\usepackage{newfloat}
\usepackage{listings}
\DeclareCaptionStyle{ruled}{labelfont=normalfont,labelsep=colon,strut=off} 
\floatstyle{ruled}
\newfloat{listing}{tb}{lst}{}
\floatname{listing}{Listing}
\title{You Only Read Once (YORO):\\Learning to Internalize Database Knowledge for Text-to-SQL}

\author {
    Hideo Kobayashi, Wuwei Lan, Peng Shi, Shuaichen Chang, \\Jiang Guo, Henghui Zhu, Zhiguo Wang, Patrick Ng
}
\affiliations {
    \textsuperscript{\rm }AWS AI\\
    \{hideodeo, lanwuwei, zhiguow, patricng\}@amazon.com
}

\usepackage{bibentry}

\begin{document}

\maketitle

\begin{abstract}
While significant progress has been made on the text-to-SQL task, recent solutions repeatedly encode the same database schema for every question,
resulting in unnecessary high
inference cost and often overlooking crucial database knowledge.
To address these issues, we propose \textbf{Y}ou \textbf{O}nly \textbf{R}ead \textbf{O}nce (YORO), a novel paradigm that directly internalizes database knowledge into the parametric knowledge of a text-to-SQL model during training
and eliminates the need for schema encoding during inference.
YORO significantly reduces the input token length by 66\%-98\%. 
Despite its shorter inputs, our empirical results demonstrate YORO's competitive performances with traditional systems on three benchmarks as well as its significant outperformance on large databases. Furthermore, YORO excels in handling questions with challenging value retrievals such as abbreviation. 
\end{abstract}

%

\section{Introduction}\label{sec:intro}

\begin{figure}[t]
  \centering
  \includegraphics[width=\columnwidth]{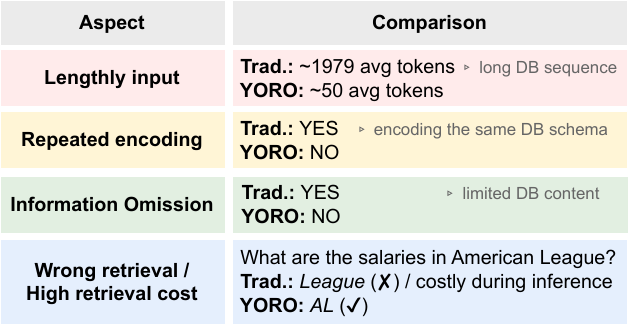}
  \caption{Comparison of traditional method and YORO.}
  \vspace{-3mm}
  \label{fig:drawbacks}
\end{figure}

The text-to-SQL task aims to convert natural language questions (NLQs) into executable SQL statements, enabling users without SQL expertise to query databases effortlessly. Existing text-to-SQL systems typically encode both a linearized database schema, which sometimes appended with partial database content \cite{lin2020bridging}, and an NLQ as input to generate a SQL query grounded in the database \cite{Scholak2021:PICARD,li2024codes}. However, this conventional approach presents several limitations, as illustrated in Figure \ref{fig:drawbacks}.

First, the repeated encoding of 
the same schema 
for every question significantly increases the computational inefficiency, especially when dealing with large database schemas. 
Second, while the linearized schema input represents the high-level structure of the database, it may still omit crucial information, such as all possible cell value choices, relationships among columns and cell values, and domain-specific knowledge.
Third, when appending schema with partial database content, existing text-to-SQL systems usually require a cell value retrieval phase for each question. This process incurs additional retrieval costs and can lead to errors if retrieval misses occur due to challenging value scenarios like abbreviations in the question \cite{chang2023dr}, resulting in incorrect SQL generation (e.g., it might fail to retrieve the value \textit{AL} given \textit{American League} in the NLQ). 

\begin{figure*}[t]
  \centering
  \includegraphics[width=2.0\columnwidth]{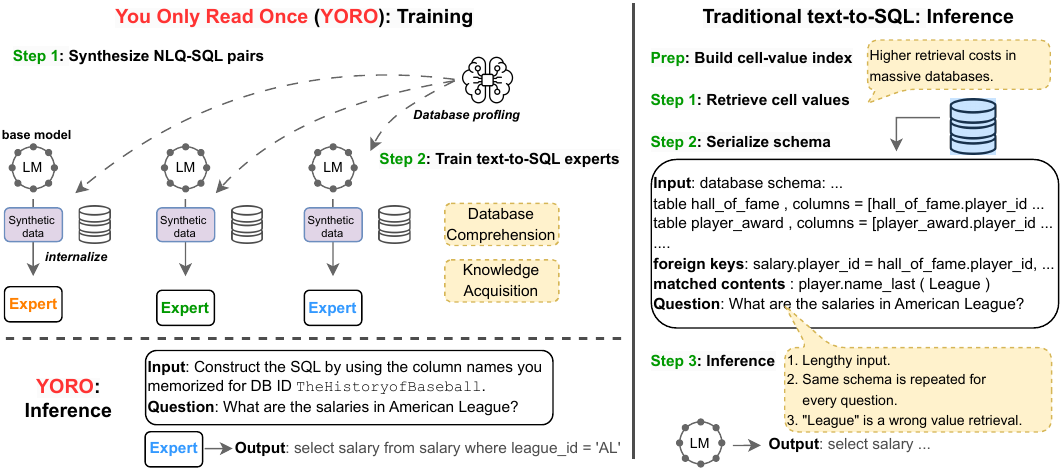}
  \caption{Overview of YORO. YORO comprehends and internalizes database knowledge through fine-tuning text-to-SQL expert models on synthetic NLQ and SQL data. Comparing with traditional methods, it leads to significantly shorter inputs and does not rely on the value retrieval step.}
  \vspace{-2mm}
  \label{fig:illustration_yoro}
\end{figure*}

To address these limitations, 
this work introduces a novel training paradigm for the text-to-SQL task, dubbed \textbf{Y}ou \textbf{O}nly \textbf{R}ead \textbf{O}nce (YORO). 
As illustrated on the left side of Figure \ref{fig:illustration_yoro},
YORO takes a fundamentally different approach by first conducting a database knowledge acquisition phase. This phase comprehensively understands the database content and directly internalizes database information into the parametric knowledge of a text-to-SQL model.
We achieve this by fine-tuning a language model on synthetic text-to-SQL data generated for target databases. 

The key advantage of YORO is evident at inference time, where it can convert NLQs into SQL queries grounded in the database without requiring schema encoding. This streamlined process contrasts sharply with the multiple steps and potential pitfalls of the conventional approach.
Furthermore, 
we propose training text-to-SQL expert models, with each expert specializing in a specific target database. 
This approach is motivated by the observation that while recent studies focus on training a single text-to-SQL model to generalize across cross-domain databases, database schemas can be highly dynamic and ambiguous with many nuances. For instance, the same column name can have different meanings in different databases.
We hypothesize that having expert models can mitigate the potential cross-database knowledge conflicts and improve overall performance.



Our extensive evaluation on popular text-to-SQL benchmarks, including Spider \cite{yu-etal-2018-spider}, KaggleDBQA \cite{lee-etal-2021-kaggledbqa}, and BIRD \cite{li2024can}, demonstrates that YORO performs competitively compared to traditional approaches across different model choices, LLaMA-7B \cite{touvron2023llama} and Mistral-7B \cite{jiang2023mistral}. Crucially, YORO achieves this performance while maintaining significantly reduced input lengths. 
Table \ref{tab:intro_stats} shows the input length comparison between YORO and two representative input formats from state-of-the-art models, PICARD \cite{Scholak2021:PICARD} and CodeS \cite{li2024codes}, across all three datasets.
YORO's input length is 66-98\% shorter than that of previous models. 
This significant reduction in input length translates to improved computational efficiency, particularly for large databases like those in the BIRD dataset, where YORO's input length remains consistent regardless of database size.
Moreover, YORO's design allows it to learn database values from synthetic data during the knowledge acquisition phase. This allows it to eliminate the separate value retrieval step during inference and learn challenging cell values. 



Our contributions are three-fold. First, we propose a novel text-to-SQL paradigm, YORO, where expert models acquire database knowledge during the training phase and utilize this knowledge to answer questions 
without having database access in inputs during inference phase.
This results in significantly shorter inputs and eliminates dependence on value retrievers.
Second, experimental results demonstrate that YORO achieves comparable performance with traditional methods. Third, our case studies reveal that YORO significantly outperforms traditional methods in large databases and excels at handling questions with challenging value retrievals. To foster further research, we will publicly release our code, synthetic data, and model artifacts.

\begin{table}[t]
  \centering
  \begingroup
  \tablefontsize
  \begin{tblr}{
  row{even} = {c},
  row{3} = {c},
  hline{1-2,5} = {-}{},
  hline{4} = {-}{dashed},
  hline{1} = {1-4}{1.0pt}, 
  hline{5} = {1-4}{1.0pt}  
}
  \textbf{Prompt}     & \textbf{Spider Dev} & \textbf{KaggleDBQA} & \textbf{BIRD Dev} \\
CodeS  & 713                 & 609                 & 1979              \\
PICARD & 137                 & 116                 & 340               \\
YORO   & \textbf{41}         & \textbf{40}         & \textbf{47}        
\end{tblr}
\endgroup
  \caption{Average input length comparison between YORO and two representative input formats from state-of-the-art models, PICARD and CodeS, across three datasets. 
  Inputs are tokenized using Mistral tokenizer.
  }
  \vspace{-3mm}
  \label{tab:intro_stats}
\end{table}

\section{Related Work}
\paragraph{Text-to-SQL.}
Fine-tuning has recently been the primary method for achieving satisfying performance for text-to-SQL~\citep{zhong2017seq2sql,yu-etal-2018-spider,Scholak2021:PICARD}.
However, with the emergence of closed-source LLMs like GPT-4 and Claude, their powerful zero-shot and in-context learning capabilities have made prompting-based solutions a strong baseline.~\citep{chen2023teaching, pourreza2024din, chang2023selective, zhang2024benchmarking, gao2023text, wang2023mac}.
For example, \citet{pourreza2024din} decomposes the parsing problem into several tasks such as schema linking, leveraging GPT-4 to solve each task by providing exemplars.
However, the effectiveness is limited by the quality of the LLMs, and it is not possible to improve the performance of these closed-source models directly.
More recently, the continued training of LLMs has been revisited 
for its potential to further boost text-to-SQL parsing performance, as seen with models like CodeLlama~\citep{roziere2023code} and CodeS~\citep{li2024codes}.
All of these methods still require schema information during inference, as well as running cell value candidate retrieval and column rankers when dealing with large databases.

\paragraph{Context Compression.}
Context compression aims to make the LLM inference more efficient by either compressing and shortening the instruction~\citep{fei2023extending,jiang2023llmlingua} or encoding the context into compact representation~\citep{chevalier2023adapting,mu2024learning,xiao2023plug}.
In this work, we aim to shorten database contents via knowledge ingestion where the contexts are "stored" in model's parameters.

Single database semantic parsing can be naturally viewed as a compressed schema setting for Text-to-SQL (e.g., ATIS~\cite{hemphill1990atis}, GEO~\cite{zelle1996learning}).
However, this approach requires a large amount of annotated examples, whereas we synthesize data. Additionally, these studies did not focus on storing database contents into model weights.
 


\paragraph{Synthesizing data for Text-to-SQL.}
Data augmentation improves text-to-SQL systems, especially in the domain generalization.  One effective approach is the skeleton-based method, which extracts SQL skeletons and populates placeholders to generate diverse SQL queries for training set's databases~\citep{zhong-etal-2020-grounded,hu2023importance}.
Subsequently, a SQL-to-Text generator such as T5 or ChatGPT is employed to produce the NLQ and SQL pairs. 
Another line of work synthesizes data for databases from new domains~\citep{wang2023dbcopilot,li2024codes}. These studies still require providing database contents as part of the input. 






\section{YORO}
YORO encompasses a training stage for database comprehension and knowledge acquisition, followed by an inference stage centered on question comprehension and SQL generation.
As illustrated in Figure \ref{fig:illustration_yoro}, we propose a straightforward approach to acquire database knowledge: synthesize a vast collection of high-quality NLQ-SQL pairs for the target database, and then continue training large language models to become Text-to-SQL experts. Once the YORO expert model is prepared, we can utilize it for online inference without accessing schema information. In contrast, traditional Text-to-SQL systems necessitate the construction of a database index and the retrieval of cell values, this time-consuming process also incurs costs for index maintenance. After cell value retrieval, we have to serialize the schema and construct a significantly longer input for model inference. In the subsequent sections, we will delve into the intricacies of our prompt design, training phase, and inference phase.




\subsection{Prompt Structure}\label{sec_prompt_struc}
\begin{figure*}[t]
  \centering
  \includegraphics[width=1.9\columnwidth]{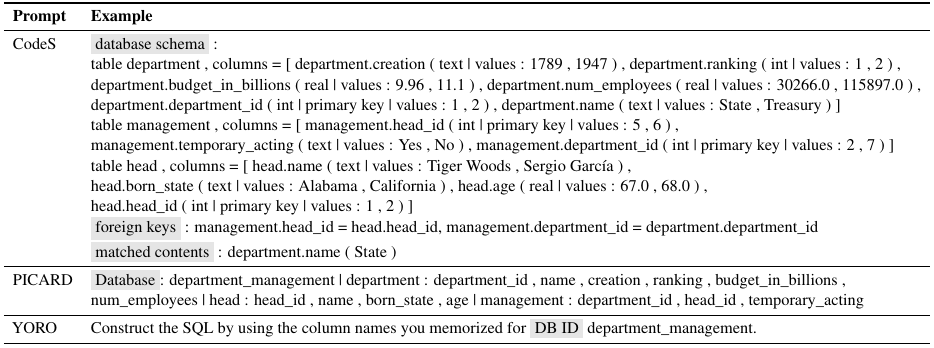}
  \caption{Examples of different prompts for the same data in Spider Dev, each followed by the same NLQ.}
  \vspace{-2mm}
  \label{tab:prompt_example}
\end{figure*}

Our prompt closely resembles the standard text-to-SQL prompts, yet it excludes all schema information (e.g., table names, column names, column aliases, column types, foreign key relationships) and cell value candidates. We only retain the database ID (e.g., department\_management). This approach prevents the model from merely copying database contents from the input when constructing a SQL query and instead compels it to internalize the database contents within the model weights for each database ID. In contrast to standard text-to-SQL prompts (Figure \ref{tab:prompt_example}), our prompt is remarkably simple and results in significantly shorter inputs.


In contrast, the CodeS prompt \citep{li2024codes} incorporates database information as extensively as possible, including table/column names, column types, sampled cell values, retrieved cell values, as well as primary and foreign key relationships. The underlying design principle is to ensure that all relevant information is accessible within the prompt. However, the PICARD prompt \citep{Scholak2021:PICARD} adopts a more simplified approach, omitting column types, sampled cell values, and foreign key relationships. Despite this, its input remains lengthy, and it adds complexity to constraint decoding during beam search.

\subsection{Database Knowledge Acquisition}\label{sec_syn_data_method}
Databases often comprise numerous tables, columns, and a vast number of rows, with each row containing relevant domain knowledge. It is a non-trivial task for a large language model (LLM) to efficiently digest and compress all this information into the model weights. More importantly, the LLM must also learn how to leverage this acquired information for the text-to-SQL task. To bridge the gap between database knowledge acquisition and text-to-SQL generation, we propose the method of continued pre-training with synthetic NLQ-SQL pairs. In this approach, we encode the database structure and cell value information within synthetic SQL queries, and then utilize these NLQ-SQL pairs for model training. This method enables the LLM to not only absorb the database information but also learn how to apply it in generating accurate SQL queries from NLQs.



In line with Zhao et al.'s \citeyearpar{hu2023importance} method, we employ skeleton-based SQL synthesis 
and condition NLQ generation based on synthetic SQLs. We leverage in-context learning with an LLM for all three stages: 1) SQL skeleton extraction, 2) SQL generation and 3) NLQ generation. We start with zero-shot prompt and gradually add generations as few-shot exemplars.
We describe each step with examples below. See Appendix A for the prompts used at these steps.

\paragraph{SQL Skeleton Extraction.} From SQL queries in a training set, we extract SQL skeletons by abstracting table names, column names, aliases, and cell values, resulting in a variety of skeletons. This diversity ensures a broad range of SQL patterns in subsequent steps, while also ensuring that the resulting synthetic SQL queries will have SQL patterns similar to those in the training set.

\begin{purplebox}
\small{\textbf{SQL:} \verb|select avg(unitprice) from track|}\\[4pt]
\small{\textbf{Skel.:} \verb|select avg(col_name) from table_name|}
\end{purplebox}

\paragraph{SQL Generation.} Using the skeletons from the previous step, we generate SQL queries for the target database by prompting an LLM to fill in placeholders within each skeleton. Multiple SQL queries are derived from each skeleton. We instruct the LLM to skip generating SQL queries for skeletons that are not applicable to the database (e.g., insufficient tables to meet the required number of tables). Each generated SQL query is executed, and those that result in execution errors are filtered out. We present the database contents in the CodeS format within the prompt, providing metadata and sample cell values for columns. We use a high temperature setting for in-context learning to ensure SQL queries with extensive coverage of different tables, columns, and cell values.


\begin{purplebox}
\small{\textbf{Skel.:} \verb|select avg(col_name) from table_name|}\\[4pt]
\small{\textbf{SQL1:} \verb|select avg(tonnage) from ship|}\\[4pt]
\small{\textbf{SQL2:} \verb|select avg(lost_in_battle) from ship|}
\end{purplebox}


\paragraph{NLQ Generation.} Finally, given a synthetic SQL from the previous step, we generate an NLQ. Interestingly, we observed in our preliminary study that in-context learning produces high-quality NLQs, whereas Zhao et al.'s \citeyearpar{hu2023importance} T5-based NLQ generator often yields unnatural NLQs, especially with complex SQL queries. In contrast, in-context learning adapts more effectively to complex queries, resulting in more natural and accurate NLQs. We believe that high-quality NLQs are crucial not only for accurate text-to-SQL learning but also for a correct understanding of the database.



\begin{purplebox}
\small{\textbf{SQL:} \verb|select avg(tonnage) from ship|}\\[4pt]
\small{\textbf{NLQ:} What is the average tonnage of the ships?}
\end{purplebox}


\subsection{Domain Experts}\label{sec_domain_exp}
YORO employs expert models to acquire knowledge for each target database. 
Unlike previous approaches that aim to train a single model capable of generalizing to unseen database schemas, our solution can be adapted to a company's proprietary databases without requiring new human-labeled data. 
Although training multiple expert models is more computationally expensive than training a single model, our solution can significantly reduce the cost of online inference. 
For example, it saves 64\% of FLOPS compared to PICARD and 93\% compared to CodeS in the 90th percentile on Spider.
Since model training is a one-time cost while inference will be continuous, YOLO will achieve better overall efficiency.

While target databases are typically unseen, the synthetic data described in Section \ref{sec_syn_data_method} profiles the target databases, allowing us to transfer domain knowledge to expert models. To train an expert model, we combine the synthetic data of the target database and the original training data (i.e., out-of-domain data), excluding its database contents, to enhance data quality and diversity. 
We hypothesize that balancing training data allows expert models to effectively mitigate cross-database knowledge conflicts. Finally, each fine-tuned expert processes the test data routed to it based on the database ID during inference.

\section{Evaluation}
\subsection{Experimental Setup}
\paragraph{Evaluation datasets.} For evaluation, we employ three widely used datasets for Text-to-SQL, Spider \cite{yu-etal-2018-spider}, KaggleDBQA \cite{lee-etal-2021-kaggledbqa}, and BIRD \cite{li2024can}\footnote{See Appendix B for statistics on these datasets.}. Spider is known to have table and column names that are simple and explicit while the other two datasets use more realistic databases containing abbreviated and ambiguous column names. BIRD's databases also contain cell values in various formats that poses new challenges for models. We use the official dev sets as our test set because our method needs to access the target database for the knowledge acquisition. 
In addition, prior studies on BIRD often use oracle knowledge during training and testing, but this is not a realistic setting. Therefore, we do not use oracle knowledge in our BIRD experiments. 


\paragraph{Evaluation metrics.} 
Text-to-SQL results are reported in execution accuracies.
We provide both micro and macro averages across all databases. Unless otherwise specified, we always present micro average results.


\paragraph{Implementation details.}
We use Anthropic's Claude-3-Sonnet model to generate synthetic data. For SQL skeleton extraction and NLQ generation, the temperature is set as 0.9 and 0.0 respectively\footnote{See Appendix C for statistics on the SQL skeletons and synthetic data.}. We set higher temperature for SQL generation since we want to obtain diverse SQLs. 

As noted in Section \ref{sec_domain_exp}, we mix the synthetic data with original training data to train each expert model. Since KaggleDBQA lacks a training set, we use the Spider's training set for KaggleDBQA experiments. We use Mistral-7B \citep{jiang2023mistral} and LLaMA-7B \citep{touvron2023llama} as the base models. We optimize these models using AdamW \citep{loshchilov2018decoupled} for 300 steps for Mistral and 500 steps for LLaMA with a batch size of 128 through gradient accumulation, a maximum learning rate of 2e-6 for Mistral and 2e-5 for LLaMA, and a linear warmup of 0.04 ratio followed by a cosine decay of the learning rate. 
The texts over 4096 tokens are trimmed during training.

\subsection{Results and Discussion}\label{sec:results_dicsussion}

\begin{table}
  \centering
  \scalebox{0.95}{
  \begingroup
\tablefontsize
  \begin{tblr}{
  cells = {c},
  cell{1}{1} = {r=2}{},
  cell{1}{2} = {c=2}{},
  cell{1}{4} = {c=2}{},
  cell{1}{6} = {c=2}{},
  cell{3}{1} = {c=7}{},
  cell{7}{1} = {c=7}{},
  hline{1,3-4,7-8,11} = {-}{},
  hline{2} = {2-7}{},
  hline{1} = {1-7}{1.0pt}, 
  hline{11} = {1-7}{1.0pt}  
}
\textbf{Method}     & \textbf{Spider Dev} &       & \textbf{KaggleDBQA} &       & \textbf{BIRD Dev} &       \\
                    & Mic.               & Mac. & Mic.               & Mac. & Mic.             & Mac. \\
\textbf{Mistral-7B} &                     &       &                     &       &                   &       \\
CodeS               & \textbf{80.2}                & \textbf{83.5}  & \textbf{44.5}                & \textbf{43.4}  & \textbf{35.7}              & \textbf{34.1}  \\
PICARD              & 76.1                & 81.3  & 37.1                & 35.3  & 22.0              & 22.1  \\
YORO                & \underline{78.5}                & \underline{81.8}  & \underline{39.0}                  & \underline{39.0}  & \underline{34.0}              & \textbf{34.1}  \\
\textbf{LLaMA-7B}   &                     &       &                     &       &                   &       \\
CodeS               & 66.9                & \underline{71.9}  & \underline{27.9}                & \underline{24.5}  & 11.7              & \underline{12.0}  \\
PICARD              & \underline{67.7}                & \underline{71.9}  & 22.8                & 22.3  & \underline{12.6}              & 11.8  \\
YORO                & \textbf{74.2}                & \textbf{76.9}  & \textbf{34.2}                & \textbf{34.3}  & \textbf{30.6}              & \textbf{30.4}  
\end{tblr}
\endgroup
}
  \caption{Performance of CodeS, PICARD, and YORO in Spider Dev, KaggleDBQA, and BIRD Dev, using LLaMA-7B and Mistral-7B. \textbf{Bold} indicates the highest accuracy, and \underline{underlined} denotes the second highest.}
  \label{tab:main_results}
  \vspace{-2mm}
\end{table}

\paragraph{Baselines.} 
We employ as our baselines models trained on the input formats of CodeS and PICARD, which contain database information as described in Section \ref{sec_prompt_struc}. We retrieved cell value candidates for the CodeS format but not for PICARD, to investigate performance differences given varying amounts of database information. This establishes two baselines: the former with full information access and the latter with minimal access. 

Results of text-to-SQL for Spider, KaggleDBQA, and BIRD are shown in Table \ref{tab:main_results}. As we can see, using Mistral as the base model consistently outperforms LLaMA. Additionally, all macro and micro average results obtained via CodeS baselines are higher than PICARD baseline results across the three datasets, except for the LLaMA versions in Spider and BIRD. This implies that various metadata and retrieved values presented in the CodeS prompt are often helpful. 
We speculate that a model might need to be powerful enough to fully utilize the rich information in the CodeS prompt and that even the LLaMA versions will perform well more consistently with CodeS prompt on all datasets if inputs are shortened via schema filtering.

\paragraph{Baselines vs YORO.} 
Results of YORO are presented in Table \ref{tab:main_results}. Our goal is to determine if a model without database access during inference can compete with or even surpass the performance of traditional methods.


First, Our method significantly outperforms all PICARD baselines by 1.9\% to 12.0\% with Mistral and 6.5\% to 18.0\% with LLaMA in terms of micro average accuracy. Although YORO's inputs lack table and column names during inference, training expert models on NLQ-SQL pairs improves performances, even surpassing PICARD baselines.

Second, when comparing YORO with CodeS baselines, we observe mixed results between using LLaMA and Mistral. With LLaMA, YORO consistently outperforms the corresponding CodeS baselines by 6.3-18.9\% in terms of micro average accuracy. This indicates that LLaMA makes it easier for YORO to answer questions using its acquired database knowledge instead of relying on the database contents in the input. On the other hand, YORO using Mistral underperforms CodeS baselines by up to 1.7-5.5\% micro average. However, a closer examination of individual database performance in Appendix D reveals that there exists several YORO experts performing as well as or better than their CodeS counterparts.

Finally, in the comparison of YORO with PICARD and CodeS in BIRD, YORO often shows either a smaller performance gap or significant improvement compared to the other datasets. This trend in BIRD can be explained by two factors. 1) the databases in BIRD are larger than those in the other two datasets, as discussed in Section \ref{sec:case_study}. 2) baselines struggle with the metadata in CodeS due to abbreviated columns and random cell values in BIRD. Our method uses synthetic NLQs to help models comprehend these abbreviated columns and cell values. This can also be seen as enhancing YORO's ability to handle complex databases by distilling the Claude's knowledge.

\begin{table}
  \centering
    \scalebox{0.95}{
  \begingroup
\tablefontsize
  \begin{tblr}{
  cells = {c},
  hlines,
  hline{1} = {1-3}{1.0pt}, 
  hline{3} = {1-3}{1.0pt}  
}
\textbf{Spider Dev (\%)} & \textbf{KaggleDBQA~(\%)} & \textbf{BIRD Dev~(\%)} \\
0.48                   & 0.37                   & 0.07                 
\end{tblr}
\endgroup
}
  \caption{Percentage of gold SQLs in each test set that exactly match any synthetic SQLs used to train our models. For BIRD, a string match is used, while for Spider and KaggleDBQA, a parsing-based match (i.e., exact match in Spider evaluation) is employed. 
  }
  \label{tab:exact_match}
  \vspace{-2mm}
\end{table}

While our discussion of these results has focused on micro average accuracy, the same trends can be observed for macro average accuracy for the most part. Overall, these results are very encouraging, especially considering that the new paradigm is a highly challenging setting\footnote{See Appendix E for error analysis.}.

To further validate our findings, we examine 
whether synthetic data significantly overlaps with test sets, causing YORO to memorize the gold labels rather than genuinely acquiring database knowledge.
Table \ref{tab:exact_match} presents the percentage of gold SQLs in each test set having an exact match with any synthetic SQLs used to train our models. It shows the ratio is less than 0.5\%, indicating that YORO is indeed learning database contents\footnote{See Appendix F for discussions on the YORO's limitations.}.

\paragraph{Different synthetic data sizes.} 
Recent studies \citep{li2023self,zhou2024lima} show the importance of quality and quantity of training data. Here, we investigate the effectiveness of scaling up the synthetic data by training models with varying amounts of synthetic data. 

\begin{figure}[t]
  \includegraphics[width=0.9\columnwidth]{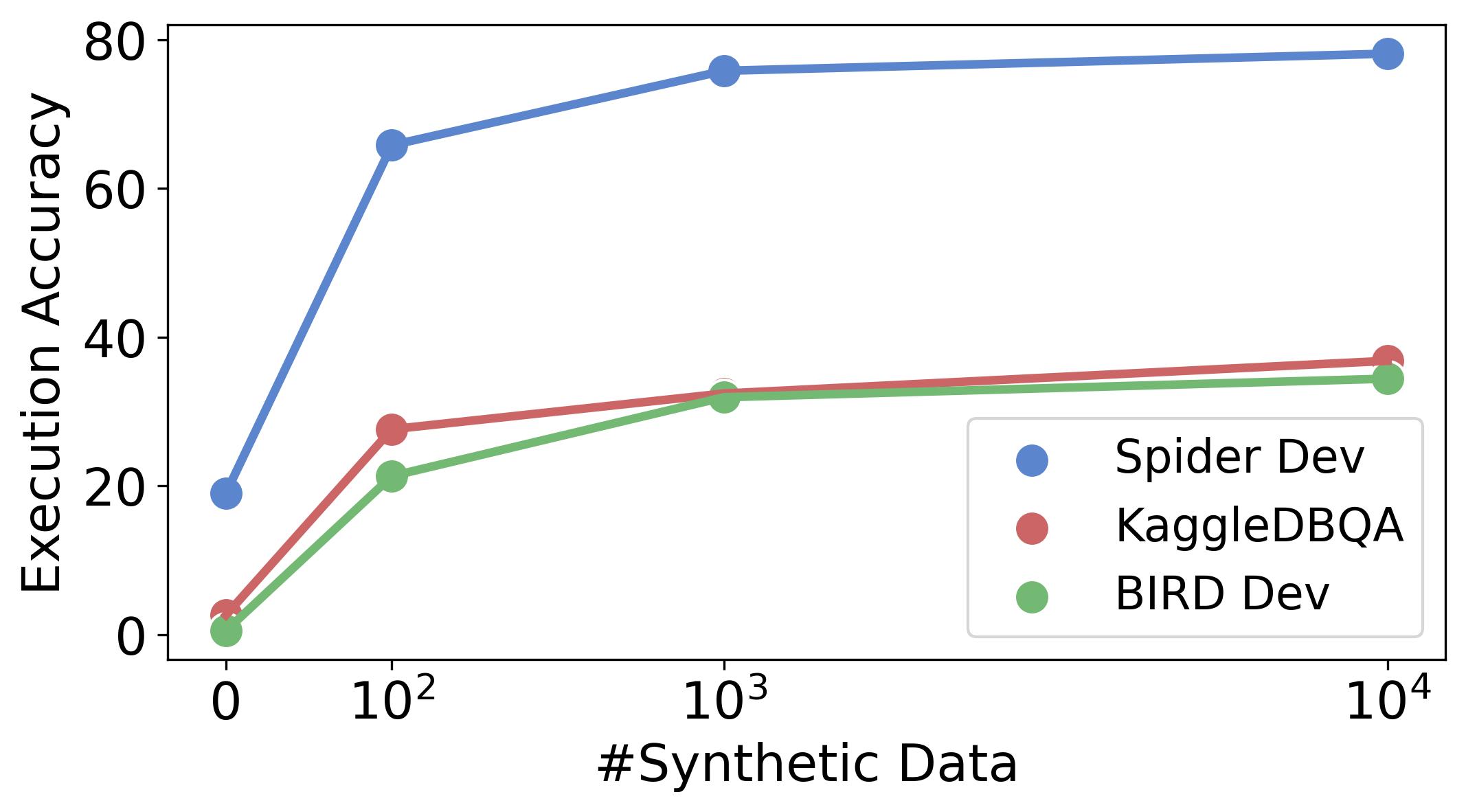}
  \caption{Performance of YORO trained with varying amounts of synthetic data using Mistral-7B.}
  \label{fig:diff_syn_size}
  \vspace{-1mm}
\end{figure}

Results of YORO with Mistral on three datasets are shown in Figure \ref{fig:diff_syn_size}, where YORO is trained on the mixed data of the original training data and different amount of synthetic data.
Mistral is a strong base model as can been seen from prior experiments. 
Without any synthetic data (i.e., using only original training data), the model achieves 0.5-19.0\% points. Surprisingly, performances improve by 20.7-46.8\% points only by being provided with one hundred synthetic data for each target database. 
We see the smaller slope after one hundred synthetic data. 
This suggests the importance of augmenting the synthetic data, although a substantial quantity may not always be necessary.

\begin{table}[t]
  \centering
  \scalebox{0.95}{
  \begingroup
\tablefontsize
  \begin{tblr}{
  cells = {c},
  cell{2}{1} = {c=4}{},
  cell{5}{1} = {c=4}{},
  hline{1-3,5-6,8} = {-}{},
  hline{1} = {1-4}{1.0pt}, 
  hline{8} = {1-4}{1.0pt}  
}
      \textbf{Method}              & \textbf{Spider Dev} & \textbf{KaggleDBQA} & \textbf{BIRD Dev} \\
\textbf{Mistral-7B} &                     &                     &                   \\
Standard            & 78.5                & 39.0                & 34.0              \\
LoRA                & 78.1                & 37.5                & 33.8              \\
\textbf{LLaMA-7B}   &                     &                     &                   \\
Standard            & 74.2                & 34.2                & 30.6              \\
LoRA                & 73.2                & 33.1                & 28.4              
\end{tblr}
\endgroup
}
  \caption{YORO with standard vs LoRA fine-tuning.}
  \label{tab:lora}
  \vspace{-3mm}
\end{table}

\paragraph{Standard fine-tuning vs LoRA.} 
We further enhance memory efficiency by employing the low-rank updating mechanism, known as LoRA \citep{hu2021lora}. 
Employing a shared foundation model with a set of LoRA adaptors for different companies could increase cost efficiency. Moreover, recent work \citep{chen2024punica} enables efficient multi-LoRA serving.
\citet{dettmers2024qlora} demonstrates LoRA's ability to compete with standard fine-tuning performances in some tasks. However, unlike typical NLP tasks, YORO requires learning domain-specific knowledge (i.e., database knowledge) that has not been seen during pre-training. The question is: would LoRA still compete with standard fine-tuning in such a new paradigm?

Results of standard fine-tuning and LoRA\footnote{We set LoRA $r=128$, $\alpha=128$ with a learning rate of 2e-4. LoRA modules are added to all linear layers of the base model following \citet{dettmers2024qlora}. All other parameters are the same as ones used for standard fine-tuning experiments.} versions of YORO with Mistral and LLaMA on three datasets are shown in Table \ref{tab:lora}. Although significantly less parameters are updated in LoRA, it lags behind the standard fine-tuning version by only 0.2-2.2\% points. This suggests that YORO is effective even with parameter-efficient fine-tuning.

\begin{table}[t]
  \centering
  \scalebox{0.95}{
  \begingroup
\tablefontsize
  \begin{tblr}{
  cells = {c},
  row{even} = {c},
  row{3} = {c},
  cell{1}{2} = {c},
  cell{1}{3} = {c},
  cell{1}{4} = {c},
  hline{1-2,5} = {-}{},
  hline{1} = {1-4}{1.0pt}, 
  hline{5} = {1-4}{1.0pt}  
}
   \textbf{Model}       & \textbf{Spider Dev} & \textbf{KaggleDBQA} & \textbf{BIRD Dev} \\
LLaMA-7B  & 73.2                & 33.1                & 28.4              \\
LLaMA-13B & 74.4                & 35.3                & 31.1              \\
LLaMA-33B & 74.9                & 36.0                  & 32.9              
\end{tblr}
\endgroup
}
  \caption{Performance of YORO across different LLaMA model sizes, all fine-tuned using LoRA.}
  \label{tab:model_size}
  \vspace{-1mm}
\end{table}

\paragraph{Different model sizes.} 
The LLaMA series encompasses a range of models differing in size.
In this experiment, we aim to investigate how varying the size of these models affects YORO's performance while utilizing LoRA fine-tuning due to computational limitations. 

Results of YORO using LLaMA with varying sizes are shown in Table \ref{tab:model_size}. 
We observe that model scaling generally leads to higher accuracies in the new paradigm. 
However, in contrast to prior work, which often shows significant improvements with increasing LLaMA model size, our results might be a more moderate enhancement.

\subsection{Ablations.} 
To evaluate the contribution of the different components in our full model, we show in Table \ref{tab:ablation} ablation results of YORO using Mistral-7B, which we obtain by removing one component at a time and retraining the model. We evaluate on holdout sets from the training data of Spider and BIRD, using 20 and 11 training databases, respectively.
\paragraph{Original training data.} First, we remove the original training data, meaning that each expert is trained solely on synthetic data. As we can see in Table \ref{tab:ablation}, Text-to-SQL accuracy drops by 5.3-7.9\% points. This indicates that it is useful to mix original training data with synthetic data.

\begin{table}[t]
  \centering
  \scalebox{0.95}{
  \begingroup
\tablefontsize
  \begin{tblr}{
  colspec = {lcc}, 
  cell{2}{1} = {c}, 
  hline{1-2,7} = {-}{black},
  hline{3} = {-}{dashed},
  hline{1} = {1-3}{1.0pt}, 
  hline{7} = {1-3}{1.0pt}  
}
                  & \textbf{Spider} & \textbf{BIRD} \\
YORO        & \textbf{74.1}       & \textbf{44.4}     \\
-- Original data  & 67.9                & 37.3              \\
-- Synthetic data & 15.2                & 0.37              \\
-- Database ID    & 71.3                & 41.3                \\
-- Domain experts & 67.5                & 40.5              
\end{tblr}
\endgroup
}
  \caption{Ablation results for YORO using Mistral-7B.}
  \label{tab:ablation}
  \vspace{-2mm}
\end{table}

\begin{figure*}[t]
  \centering
  \includegraphics[width=1.85\columnwidth]{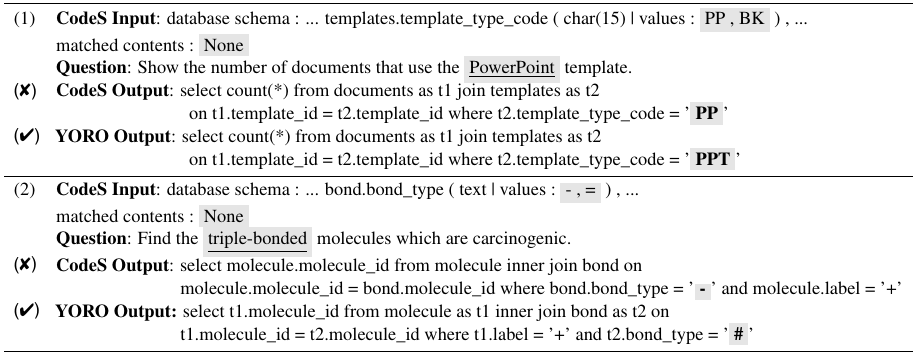}
  \caption{Examples of challenging value retrieval scenarios. The value retriever finds no value in these examples.}
  \vspace{-2mm}
  \label{tab:case_cell_val}
\end{figure*}

\paragraph{Synthetic data.} Ablating the synthetic data means that all experts are trained using only the original training data (i.e., out-of-domain data). This ablation resembles the part of the prior experiment where we trained models with different amounts of synthetic data. Comparing with the full model results, Text-to-SQL accuracy drops by 33.5-59.5\% points. This suggests the effectiveness of using our synthetic data as well as the difficulty of the new paradigm.
\paragraph{Database ID.} Next, we remove database ID from YORO's prompts during training and testing. Recall that an expert is trained on a mix of original and synthetic data, requiring the model to link questions with database knowledge from multiple sources. 
Text-to-SQL accuracy drops by 2.1-6.5\% points, showing the important role played by database ID.
\paragraph{Domain experts.} Finally, we ablate domain experts by training a single model on the mixture of synthetic data for all target databases. Text-to-SQL accuracy drops by 5.9-6.8\% points, suggesting the
positive contribution of domain experts. The expert approach helps a model to more efficiently transfer the domain knowledge for the target database through synthetic data.

\subsection{Case Studies.}\label{sec:case_study}
We explore further strengths of YORO alongside its significantly shorter inputs. 

\paragraph{Large databases.}
YORO can handle large databases consisting of many columns and tables without making input longer during inference, while traditional methods might suffer from long input sequence both in terms of accuracy and inference cost. 
To examine performances with large databases, 
we construct our evaluation set using a subset of the BIRD dev set, ensuring that each database contains at least 90 columns. This accounts for 583 of 1534 examples\footnote{The Spider dev set and KaggleDBQA do not contain such large databases.}.

Results of CodeS, PICARD, and YORO using Mistral-7B are 26.1\%, 18.4\%, and 31.6\%, respectively. 
Despite YORO underperforming CodeS on the entire BIRD dataset as shown in Table \ref{tab:main_results}, it surpasses both CodeS and PICARD when handling large databases.
This advantage over CodeS may arise in part from its input truncation when exceeding a maximum length, which is common due to CodeS's longer inputs with large databases. While schema filtering and model quantization are potential memory-saving techniques, our method does not require schema filtering during inference for handling massive databases, and it can also seamlessly integrates with model quantization.

\paragraph{Challenging value retrievals.}

In certain instances, YORO successfully generates accurate queries for questions involving challenging value retrieval. Many state-of-the-art models rely on string matching to align indicators in the question with database values, potentially leading to failures when encountering specific instances like abbreviations such as "\textbf{PPT}" in Example (1) in Figure \ref{tab:case_cell_val}. While CodeS input may occasionally include the required value among its example column values, this is not always the case. In contrast, YORO may have encountered the required value in synthetic data during training, allowing it to effectively leverage this knowledge to address such challenging cases. In addition to abbreviation, Examples (2) presents another challenging case where the retriever needs to find the database value "\textbf{\#}" from the indicator "\underline{triple-bonded}" in the question\footnote{The bonds of type "\underline{triple-bonded}" is conventionally represented as "\textbf{\#}" in many chemical notation systems.}. While the baseline model might attempt to guess the value based on the world knowledge it acquired during pre-training, this approach is not always successful. 
These examples highlight a potential advantage of YORO in handling challenging value retrieval. However, further research with a more extensive dataset is needed to validate this quantitatively.

Baselines might be enhanced by employing a more sophisticated retriever (e.g., embedding-based) with the drawback of higher retrieval costs. However, YORO's strength lies in its independence from such a retrieval step. This is particularly advantageous when dealing with massive databases containing numerous cell values. 

\section{Conclusion}
Motivated in part by recent studies on compressing instructions and contexts, this work introduced YORO, a new training paradigm for Text-to-SQL. It enables database knowledge acquisition by fine-tuning domain experts on text-to-SQL synthetic data and answers questions without having database access. YORO outperformed all PICARD baselines using both Mistral and LLaMA while also outperforming CodeS baselines when using LLaMA. 




\bibliography{main}

\begin{thebibliography}{33}
\providecommand{\natexlab}[1]{#1}

\bibitem[{Chang and Fosler-Lussier(2023)}]{chang2023selective}
Chang, S.; and Fosler-Lussier, E. 2023.
\newblock Selective demonstrations for cross-domain text-to-sql.
\newblock \emph{arXiv preprint arXiv:2310.06302}.

\bibitem[{Chang et~al.(2023)Chang, Wang, Dong, Pan, Zhu, Li, Lan, Zhang, Jiang,
  Lilien et~al.}]{chang2023dr}
Chang, S.; Wang, J.; Dong, M.; Pan, L.; Zhu, H.; Li, A.~H.; Lan, W.; Zhang, S.;
  Jiang, J.; Lilien, J.; et~al. 2023.
\newblock Dr. spider: A diagnostic evaluation benchmark towards text-to-sql
  robustness.
\newblock \emph{arXiv preprint arXiv:2301.08881}.

\bibitem[{Chen et~al.(2024)Chen, Ye, Wu, Zhuo, Ceze, and
  Krishnamurthy}]{chen2024punica}
Chen, L.; Ye, Z.; Wu, Y.; Zhuo, D.; Ceze, L.; and Krishnamurthy, A. 2024.
\newblock Punica: Multi-tenant lora serving.
\newblock \emph{Proceedings of Machine Learning and Systems}, 6: 1--13.

\bibitem[{Chen et~al.(2023)Chen, Lin, Sch{\"a}rli, and Zhou}]{chen2023teaching}
Chen, X.; Lin, M.; Sch{\"a}rli, N.; and Zhou, D. 2023.
\newblock Teaching large language models to self-debug.
\newblock \emph{arXiv preprint arXiv:2304.05128}.

\bibitem[{Chevalier et~al.(2023)Chevalier, Wettig, Ajith, and
  Chen}]{chevalier2023adapting}
Chevalier, A.; Wettig, A.; Ajith, A.; and Chen, D. 2023.
\newblock Adapting Language Models to Compress Contexts.
\newblock In \emph{Proceedings of the 2023 Conference on Empirical Methods in
  Natural Language Processing}, 3829--3846.

\bibitem[{Dettmers et~al.(2024)Dettmers, Pagnoni, Holtzman, and
  Zettlemoyer}]{dettmers2024qlora}
Dettmers, T.; Pagnoni, A.; Holtzman, A.; and Zettlemoyer, L. 2024.
\newblock Qlora: Efficient finetuning of quantized llms.
\newblock \emph{Advances in Neural Information Processing Systems}, 36.

\bibitem[{Fei et~al.(2023)Fei, Niu, Zhou, Hou, Bai, Deng, and
  Han}]{fei2023extending}
Fei, W.; Niu, X.; Zhou, P.; Hou, L.; Bai, B.; Deng, L.; and Han, W. 2023.
\newblock Extending Context Window of Large Language Models via Semantic
  Compression.
\newblock \emph{arXiv preprint arXiv:2312.09571}.

\bibitem[{Gao et~al.(2023)Gao, Wang, Li, Sun, Qian, Ding, and
  Zhou}]{gao2023text}
Gao, D.; Wang, H.; Li, Y.; Sun, X.; Qian, Y.; Ding, B.; and Zhou, J. 2023.
\newblock Text-to-sql empowered by large language models: A benchmark
  evaluation.
\newblock \emph{arXiv preprint arXiv:2308.15363}.

\bibitem[{Hemphill, Godfrey, and Doddington(1990)}]{hemphill1990atis}
Hemphill, C.~T.; Godfrey, J.~J.; and Doddington, G.~R. 1990.
\newblock The ATIS spoken language systems pilot corpus.
\newblock In \emph{Speech and Natural Language: Proceedings of a Workshop Held
  at Hidden Valley, Pennsylvania, June 24-27, 1990}.

\bibitem[{Hu et~al.(2021)Hu, Wallis, Allen-Zhu, Li, Wang, Wang, Chen
  et~al.}]{hu2021lora}
Hu, E.~J.; Wallis, P.; Allen-Zhu, Z.; Li, Y.; Wang, S.; Wang, L.; Chen, W.;
  et~al. 2021.
\newblock LoRA: Low-Rank Adaptation of Large Language Models.
\newblock In \emph{International Conference on Learning Representations}.

\bibitem[{Hu et~al.(2023)Hu, Zhao, Jiang, Lan, Zhu, Chauhan, Li, Pan, Wang,
  Hang et~al.}]{hu2023importance}
Hu, Y.; Zhao, Y.; Jiang, J.; Lan, W.; Zhu, H.; Chauhan, A.; Li, A.~H.; Pan, L.;
  Wang, J.; Hang, C.-W.; et~al. 2023.
\newblock Importance of Synthesizing High-quality Data for Text-to-SQL Parsing.
\newblock In \emph{Findings of the Association for Computational Linguistics:
  ACL 2023}, 1327--1343.

\bibitem[{Jiang et~al.(2023{\natexlab{a}})Jiang, Sablayrolles, Mensch, Bamford,
  Chaplot, Casas, Bressand, Lengyel, Lample, Saulnier
  et~al.}]{jiang2023mistral}
Jiang, A.~Q.; Sablayrolles, A.; Mensch, A.; Bamford, C.; Chaplot, D.~S.; Casas,
  D. d.~l.; Bressand, F.; Lengyel, G.; Lample, G.; Saulnier, L.; et~al.
  2023{\natexlab{a}}.
\newblock Mistral 7B.
\newblock \emph{arXiv preprint arXiv:2310.06825}.

\bibitem[{Jiang et~al.(2023{\natexlab{b}})Jiang, Wu, Lin, Yang, and
  Qiu}]{jiang2023llmlingua}
Jiang, H.; Wu, Q.; Lin, C.-Y.; Yang, Y.; and Qiu, L. 2023{\natexlab{b}}.
\newblock Llmlingua: Compressing prompts for accelerated inference of large
  language models.
\newblock \emph{arXiv preprint arXiv:2310.05736}.

\bibitem[{Lee, Polozov, and Richardson(2021)}]{lee-etal-2021-kaggledbqa}
Lee, C.-H.; Polozov, O.; and Richardson, M. 2021.
\newblock {K}aggle{DBQA}: Realistic Evaluation of Text-to-{SQL} Parsers.
\newblock In Zong, C.; Xia, F.; Li, W.; and Navigli, R., eds.,
  \emph{Proceedings of the 59th Annual Meeting of the Association for
  Computational Linguistics and the 11th International Joint Conference on
  Natural Language Processing (Volume 1: Long Papers)}, 2261--2273. Online:
  Association for Computational Linguistics.

\bibitem[{Li et~al.(2024{\natexlab{a}})Li, Zhang, Liu, Fan, Zhang, Zhu, Wei,
  Pan, Li, and Chen}]{li2024codes}
Li, H.; Zhang, J.; Liu, H.; Fan, J.; Zhang, X.; Zhu, J.; Wei, R.; Pan, H.; Li,
  C.; and Chen, H. 2024{\natexlab{a}}.
\newblock Codes: Towards building open-source language models for text-to-sql.
\newblock \emph{Proceedings of the ACM on Management of Data}, 2(3): 1--28.

\bibitem[{Li et~al.(2024{\natexlab{b}})Li, Hui, Qu, Yang, Li, Li, Wang, Qin,
  Geng, Huo et~al.}]{li2024can}
Li, J.; Hui, B.; Qu, G.; Yang, J.; Li, B.; Li, B.; Wang, B.; Qin, B.; Geng, R.;
  Huo, N.; et~al. 2024{\natexlab{b}}.
\newblock Can llm already serve as a database interface? a big bench for
  large-scale database grounded text-to-sqls.
\newblock \emph{Advances in Neural Information Processing Systems}, 36.

\bibitem[{Li et~al.(2023)Li, Yu, Zhou, Schick, Levy, Zettlemoyer, Weston, and
  Lewis}]{li2023self}
Li, X.; Yu, P.; Zhou, C.; Schick, T.; Levy, O.; Zettlemoyer, L.; Weston, J.~E.;
  and Lewis, M. 2023.
\newblock Self-Alignment with Instruction Backtranslation.
\newblock In \emph{The Twelfth International Conference on Learning
  Representations}.

\bibitem[{Lin, Socher, and Xiong(2020)}]{lin2020bridging}
Lin, X.~V.; Socher, R.; and Xiong, C. 2020.
\newblock Bridging textual and tabular data for cross-domain text-to-SQL
  semantic parsing.
\newblock \emph{arXiv preprint arXiv:2012.12627}.

\bibitem[{Loshchilov and Hutter(2018)}]{loshchilov2018decoupled}
Loshchilov, I.; and Hutter, F. 2018.
\newblock Decoupled Weight Decay Regularization.
\newblock In \emph{International Conference on Learning Representations}.

\bibitem[{Mu, Li, and Goodman(2024)}]{mu2024learning}
Mu, J.; Li, X.; and Goodman, N. 2024.
\newblock Learning to compress prompts with gist tokens.
\newblock \emph{Advances in Neural Information Processing Systems}, 36.

\bibitem[{Pourreza and Rafiei(2024)}]{pourreza2024din}
Pourreza, M.; and Rafiei, D. 2024.
\newblock Din-sql: Decomposed in-context learning of text-to-sql with
  self-correction.
\newblock \emph{Advances in Neural Information Processing Systems}, 36.

\bibitem[{Roziere et~al.(2023)Roziere, Gehring, Gloeckle, Sootla, Gat, Tan,
  Adi, Liu, Remez, Rapin et~al.}]{roziere2023code}
Roziere, B.; Gehring, J.; Gloeckle, F.; Sootla, S.; Gat, I.; Tan, X.~E.; Adi,
  Y.; Liu, J.; Remez, T.; Rapin, J.; et~al. 2023.
\newblock Code llama: Open foundation models for code.
\newblock \emph{arXiv preprint arXiv:2308.12950}.

\bibitem[{Scholak, Schucher, and Bahdanau(2021)}]{Scholak2021:PICARD}
Scholak, T.; Schucher, N.; and Bahdanau, D. 2021.
\newblock {PICARD}: Parsing Incrementally for Constrained Auto-Regressive
  Decoding from Language Models.
\newblock In \emph{Proceedings of the 2021 Conference on Empirical Methods in
  Natural Language Processing}, 9895--9901. Association for Computational
  Linguistics.

\bibitem[{Touvron et~al.(2023)Touvron, Lavril, Izacard, Martinet, Lachaux,
  Lacroix, Rozi{\`e}re, Goyal, Hambro, Azhar et~al.}]{touvron2023llama}
Touvron, H.; Lavril, T.; Izacard, G.; Martinet, X.; Lachaux, M.-A.; Lacroix,
  T.; Rozi{\`e}re, B.; Goyal, N.; Hambro, E.; Azhar, F.; et~al. 2023.
\newblock Llama: Open and efficient foundation language models.
\newblock \emph{arXiv preprint arXiv:2302.13971}.

\bibitem[{Wang et~al.(2023{\natexlab{a}})Wang, Ren, Yang, Liang, Bai, Zhang,
  Yan, and Li}]{wang2023mac}
Wang, B.; Ren, C.; Yang, J.; Liang, X.; Bai, J.; Zhang, Q.-W.; Yan, Z.; and Li,
  Z. 2023{\natexlab{a}}.
\newblock Mac-sql: Multi-agent collaboration for text-to-sql.
\newblock \emph{arXiv preprint arXiv:2312.11242}.

\bibitem[{Wang et~al.(2023{\natexlab{b}})Wang, Lin, Han, Sun, Chen, Wang, and
  Zeng}]{wang2023dbcopilot}
Wang, T.; Lin, H.; Han, X.; Sun, L.; Chen, X.; Wang, H.; and Zeng, Z.
  2023{\natexlab{b}}.
\newblock DBCopilot: Scaling Natural Language Querying to Massive Databases.
\newblock \emph{arXiv preprint arXiv:2312.03463}.

\bibitem[{Xiao et~al.(2023)Xiao, Zhang, Han, Chan, Lin, Liu, Li, Li, Cao, and
  Sun}]{xiao2023plug}
Xiao, C.; Zhang, Z.; Han, X.; Chan, C.-M.; Lin, Y.; Liu, Z.; Li, X.; Li, Z.;
  Cao, Z.; and Sun, M. 2023.
\newblock Plug-and-Play Document Modules for Pre-trained Models.
\newblock In \emph{Proceedings of the 61st Annual Meeting of the Association
  for Computational Linguistics (Volume 1: Long Papers)}, 15713--15729.

\bibitem[{Yu et~al.(2018)Yu, Zhang, Yang, Yasunaga, Wang, Li, Ma, Li, Yao,
  Roman, Zhang, and Radev}]{yu-etal-2018-spider}
Yu, T.; Zhang, R.; Yang, K.; Yasunaga, M.; Wang, D.; Li, Z.; Ma, J.; Li, I.;
  Yao, Q.; Roman, S.; Zhang, Z.; and Radev, D. 2018.
\newblock {S}pider: A Large-Scale Human-Labeled Dataset for Complex and
  Cross-Domain Semantic Parsing and Text-to-{SQL} Task.
\newblock In Riloff, E.; Chiang, D.; Hockenmaier, J.; and Tsujii, J., eds.,
  \emph{Proceedings of the 2018 Conference on Empirical Methods in Natural
  Language Processing}, 3911--3921. Brussels, Belgium: Association for
  Computational Linguistics.

\bibitem[{Zelle and Mooney(1996)}]{zelle1996learning}
Zelle, J.~M.; and Mooney, R.~J. 1996.
\newblock Learning to parse database queries using inductive logic programming.
\newblock In \emph{Proceedings of the national conference on artificial
  intelligence}, 1050--1055.

\bibitem[{Zhang et~al.(2024)Zhang, Ye, Du, Hu, Li, Yang, Liu, Zhao, Li, and
  Mao}]{zhang2024benchmarking}
Zhang, B.; Ye, Y.; Du, G.; Hu, X.; Li, Z.; Yang, S.; Liu, C.~H.; Zhao, R.; Li,
  Z.; and Mao, H. 2024.
\newblock Benchmarking the Text-to-SQL Capability of Large Language Models: A
  Comprehensive Evaluation.
\newblock \emph{arXiv preprint arXiv:2403.02951}.

\bibitem[{Zhong et~al.(2020)Zhong, Lewis, Wang, and
  Zettlemoyer}]{zhong-etal-2020-grounded}
Zhong, V.; Lewis, M.; Wang, S.~I.; and Zettlemoyer, L. 2020.
\newblock Grounded Adaptation for Zero-shot Executable Semantic Parsing.
\newblock In Webber, B.; Cohn, T.; He, Y.; and Liu, Y., eds., \emph{Proceedings
  of the 2020 Conference on Empirical Methods in Natural Language Processing
  (EMNLP)}, 6869--6882. Online: Association for Computational Linguistics.

\bibitem[{Zhong, Xiong, and Socher(2017)}]{zhong2017seq2sql}
Zhong, V.; Xiong, C.; and Socher, R. 2017.
\newblock Seq2sql: Generating structured queries from natural language using
  reinforcement learning.
\newblock \emph{arXiv preprint arXiv:1709.00103}.

\bibitem[{Zhou et~al.(2024)Zhou, Liu, Xu, Iyer, Sun, Mao, Ma, Efrat, Yu, Yu
  et~al.}]{zhou2024lima}
Zhou, C.; Liu, P.; Xu, P.; Iyer, S.; Sun, J.; Mao, Y.; Ma, X.; Efrat, A.; Yu,
  P.; Yu, L.; et~al. 2024.
\newblock Lima: Less is more for alignment.
\newblock \emph{Advances in Neural Information Processing Systems}, 36.

\end{thebibliography}

\end{document}